\documentclass[11pt]{article}
\usepackage[utf8]{inputenc}

\usepackage{natbib}

\usepackage{setspace}

\usepackage[final]{pdfpages}
\usepackage{amsmath}
\usepackage{graphicx}

\usepackage[linesnumbered,ruled]{algorithm2e}

\usepackage{hyperref,color}

\usepackage{alltt}
\usepackage{xcolor}

\usepackage{array}
\usepackage{amssymb}
\usepackage{extarrows}
\usepackage{graphicx}
\usepackage{subcaption}

\usepackage{lineno} 

\usepackage{mathtools}

\usepackage{float}
\usepackage[section]{placeins}

\newcommand{\bs}{\bigskip}
\newcommand{\bss}{\bigskip\bigskip}

\newcommand{\lrarrow}{\longrightarrow}

\renewcommand{\indent}{\hspace*{2em}}

\modulolinenumbers[5]

\usepackage{xcolor}

\usepackage{sectsty}

\sectionfont{\Large}
\subsectionfont{\large}

\usepackage{amsthm}
\theoremstyle{plain}

\theoremstyle{definition}
\newtheorem{example}{Example}

\title{Towards Coinductive Models 
for Natural Language Understanding \\ 
\normalsize{Bringing together Deep Learning and Deep Semantics}}

\author{Wlodek Zadrozny$^{1,2,}$ \\
wzadrozn@uncc.edu\\
$^1$ College of Computing, University of North Carolina at Charlotte\\
$^2$ School of Data Science, University of North Carolina at Charlotte}

\date{May-November 2020; version 1.0}

\begin{document}

\maketitle

\begin{abstract}

This article contains a proposal to add coinduction to the computational apparatus of natural language understanding. This, we argue, will provide a basis for more realistic, computationally sound, and scalable models of natural language dialogue, syntax and semantics. Given that the bottom up, inductively constructed, semantic and syntactic structures are brittle, and seemingly incapable of adequately representing the meaning of longer sentences or realistic dialogues, natural language understanding is in need of a new foundation. Coinduction, which uses top down constraints, has been successfully used in the design of operating systems and programming languages. Moreover,  implicitly it has been present in text mining, machine translation, and in some attempts to model intensionality and modalities, which provides evidence that it works. 
This article shows high level formalizations of some of such uses. 

Since coinduction and induction can coexist, they can provide a common language and a conceptual model for research in natural language understanding. In particular, such an opportunity seems to be emerging in research on compositionality. This article shows several examples of the joint appearance of induction and coinduction in natural language processing. We argue that the known individual limitations of induction and coinduction can be overcome in empirical settings by a combination of the the two methods. We see an open problem in providing a theory of their joint use.

\end{abstract}
\color{black}

\color{black}
\newpage
\begin{small}

\tableofcontents

\end{small}

\normalsize

\section{Introduction and Motivation}
\label{sec:intro}

In this article we are proposing to add \textit{coinduction}\footnote{Throughout this article we use the term `coinduction' in its most generic meaning, encompassing also coalgebra, corecursion, and bisimulation. This terminology will be explained.} to the computational apparatus of natural language semantics. This, we argue, will provide a basis for a more realistic, computationally sound, and scalable model of natural language understanding. Given that the bottom up, \textit{inductively}\footnote{We use the terms `induction' and `inductive' in their logical and mathematical sense, e.g. as in 	`definition by induction' or `proof by induction,' and not in the philosophical sense of deriving general knowledge from specific cases, as in `inductive reasoning.'}
 constructed, semantic structures are brittle, and seemingly incapable of correctly representing the meanings of longer sentences or realistic dialogues, semantics is in the need of a new foundation. Coinduction, which uses top down constraints, has been successfully used in the design of operating systems and programming languages. Moreover,  implicitly it has been present in text mining, machine translation, and in some attempts to model intensionality and modalities. So, there is scattered evidence it works. Since coinduction and induction can coexist, they can provide a common language and a conceptual model for research in natural language (NL) understanding.

We elaborate on this proposal in several ways. We motivate it by discussing the accuracy and conceptual gaps between inductive and coinductive views of NL semantics. We introduce the coinduction, coalgebras and related concepts focusing on intuitions and referring the reader to other works for in depth treatments. We show the natural match between coinduction and several natural language processing (NLP) tasks such as modeling dialogue and text mining. 
And we show examples of how induction and coinduction can jointly improve the process of assigning representations to text.  

We argue for the joint use of deep learning and deep semantics in natural language understanding. Just as the tensor product allows us to jointly explore and use two different but related algebras or vector spaces, we imagine induction and coinduction as jointly providing a better foundation for NL understanding. Although in the remainder of this article we try to convey some intuitions about their joint use, the mathematical and computational requirements for their optimal joint use are not at this point clear to us.

\subsection{Motivation}

Our motivation to pursue this topic comes from two sources, which we elaborate below. The first one is the difference in concepts used in deep learning vs. traditional semantics. The second one has to do with the limitations of processing of long sentences using the traditional semantic representation vs. the relatively successful assignment of much shallower structures using deep learning. Our proposal to think coinductively about the latter allows us to incorporate both methodologies within a single conceptual framework.

\subsubsection{Motivation \#1: The conceptual gap between deep neural networks and deep semantic analysis}

Intuitively there is a gap between using deep neural networks for natural language processing (NLP) and using deep semantic analysis for natural language understanding (NLU). If we dig deeper into this gap we might observe that their 
conceptual apparatus is different. 

\textit{Reading the textbooks. }This can be perhaps most clearly seen in the new version of a leading NLP textbook. Looking at Chapter 16 ``Logical Representations of Sentence Meaning"\footnote{We are using here the manuscript from  \url{https://web.stanford.edu/~jurafsky/slp3/}, version from  October 16, 2019} we notice it not sharing the vocabulary of the encoder-decoder and embedding models introduced earlier in the book. This is not a criticism of the book: first, this is work on progress; second, a currently missing section might create a bridge. Our point is that a bridge is needed.

To reverse the perspective, logical representations do not appear in deep learning focused NLP books such as \cite{hapke2019natural} and \cite{zhang2020dive}, and NLP doesn't appear as topic in \cite{Goodfellow-et-al-2016}.

A similar gap can be seen in \cite{nltkbook}, where lambda calculus and discourse representation is avoided in the sections mentioning the applications of logistic regression and Naive Bayes to NLP, and vice versa. 

Even much earlier the problem of bridging the two views of language, one governed by rules and the other by observations was discussed at length (e.g. \cite{klavans1996balancing}),  but arguably with little impact on the field. Somewhat similar sentiment has been expressed more recently in \cite{manning2015computational}, commenting on capabilities of deep learning: 
``really dramatic gains may only have been possible on true signal processing tasks."

This article takes the position that such bridge should be formed by creating an abstraction of both approaches, and not by an ad hoc combination. The value of this abstraction could lie in informing the theory, i.e. models of meaning, but it could also be in guiding the process of creation of better tools for human-computer interaction and natural language understanding. 

The historical analogy we might keep in mind is the creation of modern computer architectures and operating systems (e.g. \cite{auslander1981evolution}), which introduced new layers of abstraction (e.g. process streams) and new disciplines (e.g. software engineering). \\

\noindent
\textit{What about recent research?}
There was no research article on Google Scholar, as of early June 2020, mentioning ``mathematics of deep learning" and ``logical inference," although aspects of both are covered in experimental research -- ``deep learning" + ``logical inference" produces about 800 hits. Thus,  
combining logical and neural model is an active area of research. For example,  \cite{hudson2019gqa} presents a data set for question answering using both scene graphs modeling of elements present in images and challenging questions about them. In context of a different problem,  \cite{zhang2019optimizing} discuss ensuring factual correctness of summaries, using two models, one logical (to attend to facts) and one neural (for reducing the size of the document). In our third example, \cite{richardson2020probing} show that neural attention based models such as BERT \citep{devlin2018bert} can be retrained to master aspects of natural language inference. On the other hand, the examples we present in Section \ref{sec:notEvery} show that deep neural networks still seem incapable of deeper reasoning without special purpose architectures, and even modeling elementary arithmetical operations is a challenge.

\subsubsection{Motivation \#2: Accuracy gap for long sentences between deep learning and deep semantic models}\label{sec:longsents}

There is a gap in the accuracy between deep neural networks and deep semantic analysis, irrespective of the fact that they try to address different aspects of natural language understanding. 

\begin{table}[]
\centering
\begin{tabular}{|l|l|}
\hline
``\textbf{Automata}" & ``\textbf{Structures}" \\
\hline
POS Tagging                                      & Parsing                       \\ \hline
Text Mining (e.g. \texttt{Person}, \texttt{Date}, ... ) &         Computational Semantics                      \\ \hline
Google Search                                    & Computational Pragmatics      \\ \hline
Gist Translation                                 & Good Translation              \\ \hline
\end{tabular}%
\caption{On the left, one intuitive computational model is that of an automaton, with data coming as infinite streams (esp. for training). 
On the right the models are assembled into structures (often by hand) from previously defined components. The accuracy of the models on the left is higher. However, the automata-based approach has problems dealing with task requiring deeper inference.}
\label{tab:4suc4fail}
\end{table}


Table \ref{tab:4suc4fail}, \textit{viewed through the lenses of the systems' ability to successfully attend to long sentences}, shows in the left column intuitively `successful' NLP applications; and in the right the areas where in our view we have seen limited progress in the last 30 years. Obviously, metrics used by the applications mentioned in the two columns are different. For example, one can argue that computational pragmatics did not exist 30 years ago, and only recently we have started to see computational, probabilistic models of pragmatics  \citep{Probabilisticpragmatics} 
\citep{scontras2017probabilistic},\footnote{\url{https://michael-franke.github.io/probLang/}
\url{http://www.mit.edu/~tessler/short-courses/2017-computational-pragmatics/} last retrieved on May 13 2020}, which suggests a big jump. 
Nevertheless, the areas on the right do not scale with sentence length. And later, in Sections  \ref{sec:coindinformly} and  \ref{sec:nliscoind}, we will argue, from a more abstract perspective, that the differences between the columns can be attributed to the differences in their respective computational models.

\FloatBarrier

Let us discuss long sentences. Prior research in this area shows how parsing accuracy decreases with the length of the sentence. For example, 
\cite{mcdonald2007characterizing} observe fast drop  in precision and recall of dependency parsing with the increase of the dependency length, the distance to the root, and length of sentences. Similar results appear in Fig.4 of
\cite{choi2015depends}.  Actually the situation might be worse than these sources suggests. In an analysis of parsing of sentences up to the length of 156,
\cite{boullier2005efficient}	  
	   entertain a possibility that “(full) parsing of long sentences would be intractable." Clearly, deep neural networks improved the accuracy of parsing. However, even with the attention-based models, \cite{nivre2020}(Fig.4) reports a $\sim$20\%  drop in the labeled attachment scores when the dependency length increases from 1 to 5.

This is clearly a problem, even for linguistically oriented data sets. A recent statistics  given in \cite{borbely2019sentence} shows that depending on the language and the corpus the average sentence length is 19-38 words. However, thousands of  sentences in each corpus are longer than 100 words. The average sentence length in the Penn Treebank is 20.54 words (and the standard deviation of 8.6); in Genesis, 34 words; but, per classic \cite{yule1939sentence}, in ``Biographia Literaria"   10\% of sentences are long and have the average length of about 70 words.

\begin{figure}[t]
\centering
\includegraphics[scale=0.7]{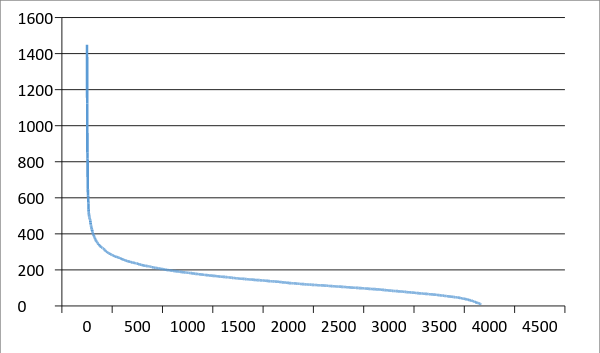}
\caption{Distribution of Claim 1 sentences lengths in 4000 patents related to sustainability. Note that over 90\% of sentences are longer than 50 words. (From \cite{rajshekhar2016analytics}).}
\label{fig:claimslength}
\end{figure}

The situation is even more problematic when we switch from general corpora to specific ones. In our previous work \citep{rajshekhar2016analytics}, we discussed the problem of parsing long sentences in the context of legal text corpora, namely patents. 
Fig. \ref{fig:claimslength} (op.cit.) shows the distribution of the lengths of the main patent claim (Claim 1). These claims are expressed as single long sentences. The sentences of Claim 1 average 150-180, and can be up to 1400 words long (different weekly data introduce the variation). Although the extreme length is due to legal rules, nevertheless we note that 93\% of the claims in this series are longer than 50 words. 	   
	   This means that any analysis of an average Claim 1 is likely to be wrong. (In our unreported experiments in 2018 on a handful of claims about 30 words long, using different dependency parsers, we did not get even a single correct parse).

\begin{figure}[t]
\centering
\includegraphics[width=0.6\textwidth]{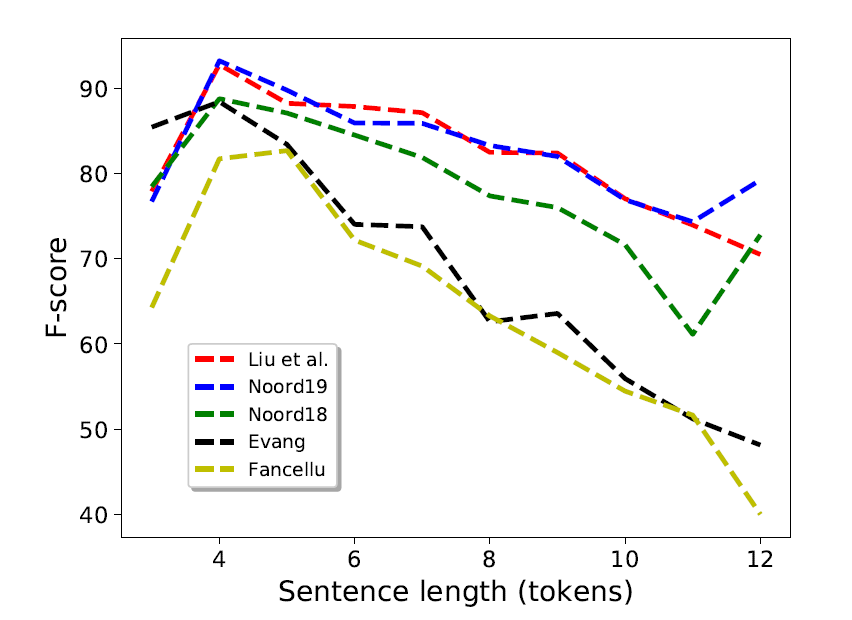}
\caption{Semantic representations are difficult to build even for short sentences, as shown by \cite{abzianidze2020first}. All systems report drop in performance with the sentence length.}
\label{fig:claspSent}
\end{figure}

What about semantic parsing? Semantic representations are difficult to build even for short sentences, as shown in Fig. \ref{fig:claspSent}, from  \cite{abzianidze2020first}. All systems submitted to the competition on the shared task on semantic parsing show  a drop in accuracy as the length of the sentence increases.

\FloatBarrier

\bss

\color{black}
\noindent
\subsection{Hypothesis}

\textit{Adding coinduction to semantics can provide a foundation 
for a more realistic, computationally sound and scalable model of natural language understanding.} \\

We are proposing adding coinduction to the computational apparatus of semantics. This we argue, will provide a basis for a more realistic, computationally sound and scalable model of natural language understanding. Given that the bottom up, inductively constructed semantic structures are brittle, and seemingly incapable of representing longer sentences or realistic dialogues, semantics is in the need of a new foundation. Coinduction, which uses top down constraints, has been successful in the design of operating systems and programming languages. Moreover, one can argue that implicitly it has been present in text mining, machine translation and in some attempts to model intensionality (even though the term itself does not appear in any articles on \texttt{aclweb.org}). 
In \cite{barwise1996vicious}, which is a good introductory textbook, it is used to describe self reference, paradoxes and modal logics.

So, there is scattered evidence coinduction works. Since coinduction and induction can coexist, they can provide a common language and conceptual model for research in NL understanding. 

We should mention that one of the first theoretical proposals to look at agent interaction as coinduction appeared in \cite{wegner1999coinductive}, and it included explicit mention of NL dialogue and question answering. Within the following twenty years, as argued in the present article, the focus of NLP shifted towards coinductive 
methods\footnote{
This shift occurred without ever mentioning the concept itself. There are literally 12 entries mentioning "deep learning" and "coinduction", mostly accidentally, although  \cite{elliott2019generalized} (a formal analysis of convolutions) 
is an exception. (As of October 2020).}, 
namely deep learning, with the theoretical justification coming from the universal approximation properties of neural networks. 
This article summarizes some of these developments and argues for an explicit introduction of the term `coinduction' to the vocabulary of NLP. \\

\noindent
To make the argument, we will focus on the following questions: 
\begin{enumerate}
\item What is coinduction? 

That is, we will discuss coinductive data, coinductive functions, coalgebras and 
coinductive proofs (only marginally). See Sections \ref{sec:coindinformly} and \ref{sec:nliscoind}.

\item Do we need it? 

We have already started to argue that we do. But we'll expand on it below in 
Sections \ref{sec:colaexam} and \ref{sec:notEvery}, where the main point will be the advantages of combining inductive and coinductive information, and contrast it with the inadequacies of only using a single approach. 
The empirical arguments presented there are based on results taken from the work of other researchers and recast in the language of coinduction.

\item What can be our next steps? 

In Section \ref{sec:summary} we hypothesize that the practical and theoretical importance of the joint inductive-coinductive approach to natural language understanding will most likely be seen in models of compositionality. 

\end{enumerate}

\color{black}
\section{What is coinduction? Informally.}\label{sec:coindinformly}

Although coinduction and coalgebra do not appear in the \texttt{aclweb.org} repository of NLP articles.  We will argue that there is already quite a bit of research in NLU  \emph{}{and in semantics} done in the co-inductive style;  it \textit{just hasn't been \textbf{{named}}   `co-inductive'}. We want to argue 
the ``classical"-style formal semantics can be extended with new problems to bridge the gap between ``coinductive" and ``inductive" views of the data.

Before we attempt to answer the question ``what is coinduction," we want to informally discuss what this speculative paper is about? Namely, in theoretical computer science we have the concept of \textit{co}-induction, or coinduction {(which is the spelling we will use). }
Induction builds structures bottom up and can be viewed as a reductionist process,\footnote{Objects are reduced to their parts.} while
coinduction provides top down constraints. Our main idea is to base NLU  \textit{partly} on coinduction, for example to provide better models of dialogue and to address the problem of parsing and understanding of long sentences.
We believe, coinduction can be incorporated into NL semantics, helped by the fact induction-coinduction relationships are relatively well investigated in formal logic and theoretical computer science.
Empirical evidence suggests we will be more successful 
in addressing difficult problems. \\

\noindent
\textbf{Five questions about coinduction we need to answer}

\begin{itemize}
\item What is coinduction? 
\item How might coinduction  be applicable to semantics? 
\item Has coinduction been applied to formal semantics of NL? 
\item  What are the limitations of coinductive view of NLP?
\item  What kind of problems might be amenable to progress using a joint induction-coinduction approach? 
\end{itemize}

This exposition will be a bit less formal than in a properly technical article, but it's good to start somewhere, and we'll provide complementary references explaining the logic(s) of coinduction. As we said earlier, applying coinduction to natural language processing is an open problem, and this article only intends to show likely places of intersection between the two fields.

\subsection{Coinductive data, coinductive functions, coinductive proofs}

We will use "coinduction" in the most generic meaning, to mean the three aspects of coinduction: 
\begin{itemize}
\item coinductive data

\item coinductive functions

\item coinductive inference: models and proofs

\end{itemize}
\bs
We will mostly focus on the first two bullets. We will not be breaking any new grounds here, either in having any insights or in arranging the material; we simply will reuse well known examples. These examples will later be used as formal models for various natural language phenomena, such as turn taking in modeling NL dialogue (Section \ref{sec:dialogue}. 

To develop computational intuitions about coinduction, we start with two very generic and known examples from programming, appearing e.g. in Python, Haskell and Prolog. The examples introduce the concepts of \textit{codata}, \textit{coinductive program},  \textit{constructor} and \textit{destructor}, see e.g.  \cite{gordoncorecursion}.

\noindent

\begin{example}\label{ex:inflist}
\textbf{Coinductive data: two styles of \texttt{list} }

\bs
\noindent
\textbf{Data:} The 4-element finite list $L4 = [1, 1, 1, 1]$ is built by specifying
$$L4 = cons(1, cons(1, cons(1, cons(1, nil))))$$ 
where $cons$ is a list constructor
and $nil$, the empty list, a nullary \textit{constructor}.

\bs

\noindent
\textbf{Codata:} The infinite list $$L1 = [1, 1, 1, 1, . . . ]$$ is defined by specifying
$hd(L1)=1$ and $tl(L1)=L1$, where $hd$ and $tl$ are \textit{destructors}.

\end{example}

\begin{example}\label{ex:addone}

\textbf{Coinductive programs: two styles of \texttt{addOne} }
 
\bs
\noindent
\textbf{Recursion} (finite lists, step by step):

\indent\texttt{AddOne(nil) = nil} 

\indent\texttt{AddOne(cons(n, l)) = cons(n+1, AddOne(l))}

\bss

\noindent
\textbf{Corecursion} (infinite lists with lazy evaluation):

\indent\texttt{null(AddOne(l))=(l=nil)} 

\indent\texttt{hd(AddOne(l))=hd(l)+1} 

\indent\texttt{tl(AddOne(l))=AddOne(tl(l))}

\bss
 
The recursively defined \texttt{AddOne}
maps finite lists to finite lists; the corecursively defined \texttt{AddOne}
maps finite lists to finite lists and infinite lists to infinite lists. 
\bs
\end{example}

More generally, recursion defines a function mapping values from a datatype by invoking itself on the components of the \textit{constructors} used to build data values. Corecursion
defines a function mapping to a codatatype by specifying the results of applying
\textit{destructors} to the results of the function.
\bigskip

\subsection{What are coalgebras}

We will introduce the concept of coalgebra informally, through a few examples, and we will mostly follow the exposition from \cite{jacobs2017introduction}, \cite{jacobs2011Slides} and \cite{rutten2000universal}. We will illustrate the concepts using well known examples from natural language processing (NLP).

\subsubsection{Algebras describe constructions}\label{sec:consvobs1}

We are assuming the reader is familiar with the concepts of an \textit{algebra}. An algebra, for example, an algebra of sets, a Boolean algebra, a group or a vector space, is defined by specifying its domain and its operations. Well known examples include set union and set difference;  conjunction, disjunction and negation for Boolean algebras; multiplication and inverse for groups; and vector addition for vector spaces. In all these cases, the operations construct new elements from the elements of the algebra. 
Pictorially, algebras with a "carrier" $X$ are maps into $X$ from some type of system or structure or an expression containing $X$ (shown as a box below).

\begin{center}

\large
\fbox{ $X \ ... \ X$} $\overset{c}\lrarrow \ X$
\normalsize
\end{center}

\begin{figure}
\centering

\includegraphics[scale=0.45]{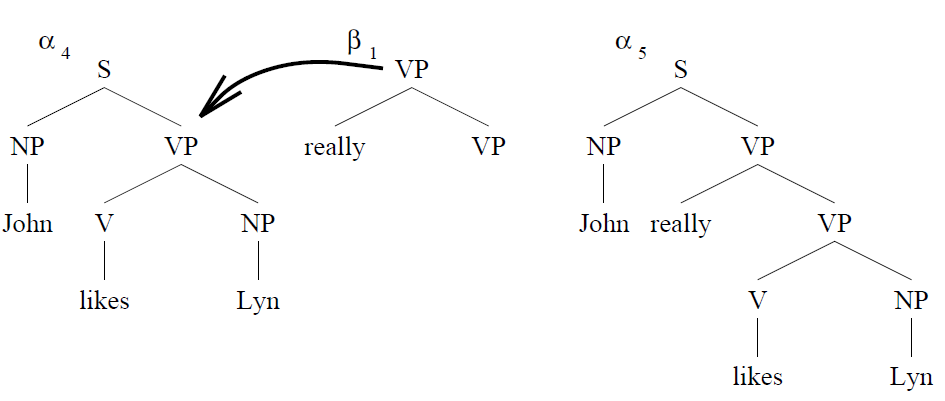}
\caption{A grammar can be viewed as an algebra on trees, where two trees can be combined through adjunction, pictured here, or substitution (not shown). The figure is reproduced from \cite{joshi2003formalism}, Figure 3.}
\label{fig:TAG}
\end{figure}

\FloatBarrier

In NLP, we can view a grammar as an algebra defining how smaller parse trees can be combined into bigger trees; perhaps this is best shown in the case of tree adjoining grammars (TAGs), in Fig.\ref{fig:TAG}, where the operation of adjunction $a$ of pairs of trees can be represented as 

\begin{center}

\large
\fbox{ $Trees \ \times \ Trees$} $\overset{a}\lrarrow \ Trees$
\normalsize
\end{center}

\noindent
or simply $a: \ Trees  \times  Trees \lrarrow \ Trees$. \ This representation focuses only on types in the domain and codomain (range) of the function $a$, and we will see in a moment that in coinduction, which we will use to represent named entity recognition (NER) in NLP,  we simply reverse the arrow. Notice that this representation tells us nothing about any constraints that the pair of trees has to satisfy for $a$ being applicable.
Such constraints have to be specified separately, as they usually are.

\subsubsection{Coalgebras describe observations}\label{sec:consvobs2}

If we were to create an "algebraic" view of the standard named entity extraction operation, we would need to put the box on the other side.

\begin{center}
\large

$X$  $\overset{c}\lrarrow$ \fbox{ $X \ ... \ X$}

\normalsize
\end{center}

\begin{figure}
\centering
\includegraphics[scale=0.36]{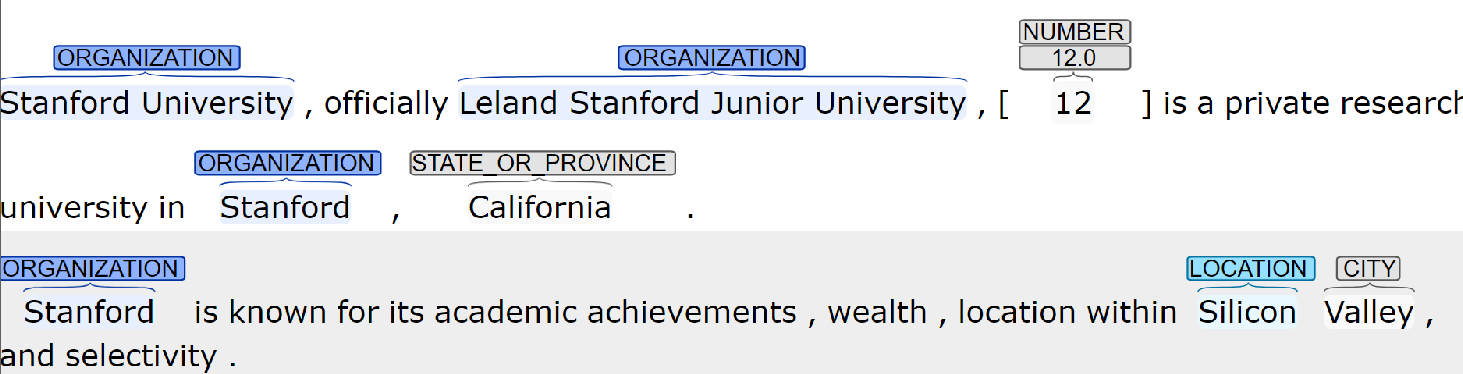}
\caption{Named entity recognition 
adds annotations to the text. In this example, the semantic types \texttt{\textsc{organization, ....}} are added to a fragment of a Wikipedia article. Note the imperfections. (We used \url{http://corenlp.run/} Stanford CoreNLP 4.0.0, updated 2020-04-16, to obtain these annotations, but such annotations are never perfect).}\label{fig:ner}

\end{figure}

\FloatBarrier

For example, let us take $X = Sentences$ to consist of lists of words produced by a sentence finding algorithm on a corpus of text data. Finding named entities or semantics types in the sentences, e.g. \texttt{person, city, geopolitical entity, cardinal} etc., produces new lists of words together with \texttt{IOB} tags, indicating positions of the substring with specific properties 
and these tags, as shown below in formula (\ref{eq:ner}), appear instead of the "..." ellipsis.

\begin{table}[]
\centering
\footnotesize
\begin{tabular}{|l|l|l|l|l|l|}
\hline
officially & Leland & Stanford & Junior & University & {[}12{]} \\ \hline
O          & B      & I        & I      & I          & O        \\ \hline
           & \multicolumn{4}{l|}{ORGANIZATION}       &          \\ \hline
\end{tabular}%
\caption{ Looking at the details of annotations in Fig.\ref{fig:ner}, the 
\texttt{IOB} annotations show the beginning \texttt{B}, inside \texttt{I} and outside \texttt{O} of the entity.}
\label{tab:IOB}
\end{table}

As shown in Figure \ref{fig:ner} and Table \ref{tab:IOB}, finding entities of interest in this model consists in \textit{adding} information to the original, that is creating a table with observations.

\color{black}

Notice we are not \textit{constructing} new strings, we are \textit{observing} the text and finding items of interest, i.e. named entities, and we are adding annotations to the text. This change of perspective creates a \textit{co-algebraic} view of the text. Notice the difference: sets of trees were combined into new trees in the previous example; however, for NER we transform sentences into \textit{expressions} containing the words of the original sentences and the IOB annotations. \\

Replacing the box by the actual types involved in producing the annotations we get, for the simple case of looking for the \texttt{person} type:

\begin{equation}\label{eq:ner}
(Words)^{<\omega} \xrightarrow{\textsc{ner:city}}  (Words \times \{I,O,B\})^{<\omega}
\end{equation}
That is, \textsc{ner:city} takes finite lists of words and produces finite lists of words with the \texttt{I,O,B} labels. The specification of a more complex NER task will be more complex, but our point here is that we are not building new structures but annotating existing entities; i.e. we are not constructing but observing. 

\begin{figure}
\centering
\includegraphics[scale=0.42]{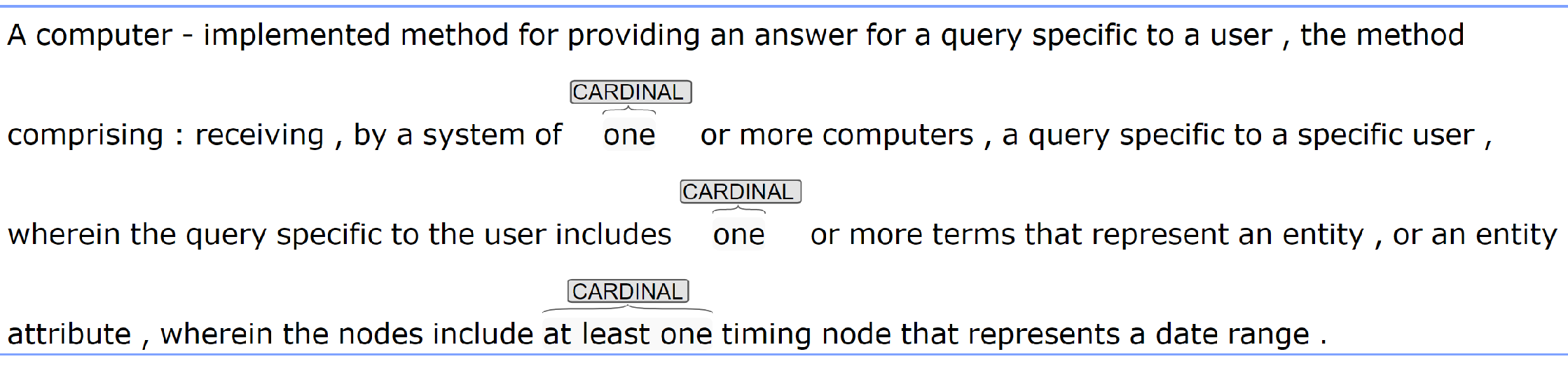}
\caption{Named entity recognition or semantic type recognition 
adds annotations to the text. In this example, the semantic type \texttt{cardinal} is added to a fragment of a patent claim. Note that although ``at least one" is annotated properly, ``one or more" is not. (We used \url{http://stanza.run/} to obtain these annotations in June 2020).}\label{fig:cardinal}  
\end{figure}

Also, as shown in Fig. \ref{fig:cardinal}, observations can be incomplete or contain errors. The point of using coalgebraic (or coinductive) specifications is to make analysis resilient to errors, and produce partial useful output.

\FloatBarrier

\subsection{A few words about related concepts}

Throughout this article we are using the words ``coinduction" and ``coinductive" in their  widest meaning, because we see it as name for a point of view 
and a generic computational approach\footnote{An alternative could be to use the words "coalgebra" and "coalgebraic"}.
We will also use these terms in a narrower technical sense referring to a specification of 
how data can be broken down into simpler items or annotations added to them by observation. There is also another narrower meaning of these words, referring to 
a proof technique, which is a ``dual" of familiar proofs by induction -- we will not concern ourselves with this specific sense. 

There are a few other related concepts which are mathematically important, but here we only want to acknowledge their existence, and perhaps provide additional intuitions. 

\FloatBarrier
For the readers familiar with logic programming it might be helpful to think in terms of 
\textit{largest vs. smallest models}.

\begin{quotation}
When we give an \textit{inductive} definition, we mean the smallest set that satisfies the given constraints; everything that's in the set has some justification. We build \textit{the smallest model (i.e. the smallest fixpoint of the construction).}
\textit{Coinductive} definitions specify the largest set that is consistent with them. 
The construction of a model proceeds by finding \textit{the largest fixpoint consistent with the specifications}\footnote{This particular reference might be helpful for the reader struggling with intuitions: 
\url{https://ask.metafilter.com/42858/What-the-heck-is-coinduction}
}.

\end{quotation}

A question worth asking is: which  of the two constructions  will be more robust in real applications? It is our intuition that largest models might tolerate better noise and novel data, including new expressions and errors in text. Usually, irrelevant things are consistent with specification, so with coinductive constructions, we do not have to predict in advance all possible cases.

For other potentially relevant concepts such as \textit{corecursion} and \textit{bisimulation} (which we are not using in this article), as well as in depth treatment of fixpoints and coinduction, we would like to refer the reader to \cite{barwise1996vicious}.

In the next section we will informally introduce \textit{process algebras}, in the context of multimodal dialogue, and informally describe their connection to coalgebras. 

\color{black}
\section{Natural language processing is intuitively coinductive}\label{sec:nliscoind}

We saw a moment ago that a formal definition of named entity recognition, a common NLP task is coinductive in its form. Now we will go through a list of standard NLP tasks and talk about them using the jargon of coinduction. The point of this exercise is twofold: to get used to coinductive definitions and to show that there is a common thread to solving natural language processing and understanding problems that require abandoning the standard inductive, build step-by-step construction of meaning.

\subsection{Automata as a coalgebra}\label{sec:autascoalg}

Finite state automata are widely used in NLP, as shown in the list and discussion below. They are naturally described in the language of coalgebras: 

\noindent
\begin{equation}\label{eq:automaton}
c : S \rightarrow \{halt\} \cup (A \times S)
\end{equation} 
Here, $S$ is a set of states, $c$ is a transition function, and $A$ are outputs (or observables). 
Note that this formula does not constrain us to finite state automata, that is this specification is more general. \\

\noindent
\textbf{Examples of "automata" used in NLP with coinductive representations:} 
\begin{itemize}
\item Any \texttt{regex}
\item Chunkers that can operate with limited information; e.g. sentence splitters, phrase finders, ","-finders, part of speech taggers,  etc.
\item Binary neural networks  
\item RNNs (LSTMs, BERT encoders) 
\end{itemize}

With some modification of the above transition function, and making the outputs $A$ more specific, along the lines of formula (\ref{eq:ner}), we can intuitively see the coinductive character of the first two items on the list. The other two items are perhaps less intuitive. However, binary neural networks are finite state automata (see \cite{vsima2020analog} for a more in depth recent discussion); recurrent neural networks (RNNs) 
share the same spirit but would have more complex definitions (see e.g. \cite{carlsson2020topological} for a mathematical description of feed forward NNs).  
Actually, \cite{sprunger2019differential} show that learning in artificial neural networks is coinductive (at least for some types of RNNs). The above list covers perhaps the majority of techniques used in NLP. Moreover it exemplifies the techniques used to address the NLP problems in the left hand side column of Table \ref{tab:4suc4fail}.
However, we want more, that is an argument that coinduction provides a natural description of several unaddressed problems in language understanding, and in particular the right hand side of the same table. 

\color{black}
\subsection{Dialogue: why it matters that interaction is coinductive}\label{sec:dialogue}

A conversation can be viewed as an unbounded sequence of turns modifying the internal states of the interlocutors. Note that only the exchanged words are observable, but obviously conversations change us. Some conversations are finite (e.g. hotel reservations), but some are "infinite", for example between members of the same family. Even our Google searches are better viewed as infinite conversations, where our past interactions and current states (e.g. location) together with the query produce a list of results but also modify the state of the Google search engine (which cannot be observed).\\

\noindent
\textbf{Dialogue coinductively}. Natural language dialogue can be described as a very simple coinductive process consisting of a finite or infinite collection of utterances, using the automaton equation (\ref{eq:automaton}).

\begin{equation}\label{eq:dialog0}
  \texttt{turn:} \ \texttt{States} \longrightarrow   
\{halt\} \cup \ (\texttt{Who} \times Words^{<\omega} \times \texttt{States})  
\end{equation}
\texttt{States} are not observable and are the internal states of interlocutors, labeled by \texttt{Who}. Labels and utterances can be observed. \\

This model is perhaps too simplistic. However it is flexible enough to capture both recent attention-based neural network models of dialogue (\citep{budzianowski2019hello}) and recent  challenges to standard views of dialogue in linguistics (\cite{gregoromichelaki2020actionism}). 

This correspondence 
deserves a longer discussion. In a "classical" view of NL dialogue (e.g. \cite{nltkbook}, Chapter 1), each utterance is a sentence or full phrase that can be assigned linguistic meaning independently of other utterances. This conceptual model admits exceptions, e.g. fragments that had to be interpreted using other, context driven mechanisms. A more recent view \cite{gregoromichelaki2020actionism} postulates a model in which a single sentence structures can be emerging across participants, and where 
speaker and hearer can exchange roles across all syntactic and semantic dependencies:

\begin{quote}

Ruth: \textit{I’m afraid I burned the kitchen ceiling.}

\vspace{-5pt}
Michael: \textit{Did you burn}

\vspace{-5pt}
Ruth: \textit{myself? No, fortunately not.}

\bs
A: \textit{Have all the students handed in}

\vspace{-5pt}
B: \textit{their term papers?}

\vspace{-5pt}
A: \textit{or even any assignments?}

\end{quote}

Obviously, formula (\ref{eq:dialog0}) is a good high level model of such exchanges. But it needs to be augmented to correctly model the linguistic acceptability of the above and the ungrammatical nature of e.g. \textit{any even assignments or?}. A general linguistically motivated method of doing so, given by \cite{gregoromichelaki2020actionism}, is based on "dynamic syntax", \cite{kempson2000dynamic}. In a very practical setting of task oriented dialogues, our earlier work, \cite{zadrozny2000natural,zadrozny1998conversation}, handled similar fragments by finding most likely semantic representations, derived from an implementation of a construction grammar (\cite{goldberg1995constructions,zadrozny1994nl,zadrozny1995significance}). Thus both theory and practice seem support this abstraction. 

It is our view that for practical applications, and, in particular, to be able to correctly assign semantic structures to long sentences, as discussed earlier in Section \ref{sec:longsents}, a combination of coinductive and syntactic constraints is necessary.
The exact proportion and formal role of each is an open issue. 
It might be the case though that they would have complementary origins, where the coinductive constraints are learned from observations using machine learning mechanisms, and the grammatical constraints would come from human designed formal grammars. 
We will return to this later in Section \ref{sec:colaexam}, providing examples of coinduction and induction working together and in Section \ref{sec:notEvery}  to argue that not everything should be done by coinduction.

\color{black}
\subsection{Multimodal interaction coinductively}

A natural extension of spoken dialogue is multimodal dialogue between people, involving both speech and gesture. More broadly multimodal interaction can involve humans and machines communicating using speech, handwriting, hand gesture and gaze.

Interestingly, there is already an explicit use of coinduction to account for speech-gesture interaction
\cite{rieser2017process}, 
even though the term "coinduction" does not appear there. However, the term ``process algebra" is in the title, and the particular version of process algebra, the $\psi$-calculi are closely related to coinduction (see below). 
The cited article argues that to account for split utterances, in the spirit of examples shown in Section \ref{sec:dialogue}, and to account for the asynchronous multimodal communication and coordination, a better model is required than ``naive compositional" models canonically employed in NL semantics. Namely, a model in which we have:

--- channels on which information (data,
agents or procedures) can be sent;

--- procedures operating concurrently;

--- interfaces enabling communication
among processes;

--- active and non-active processes;

--- communication among agents
organized via an i-o-mechanism, and where 

--- ``composition does play a role finally, when the speech-gesture contact points have been identified."

These are natural postulates in context of multimodal interaction. Semantically, they signify a transition from a static model of semantics to a dynamic one. This is similar in spirit to the ``dynamic syntax" approach to dialogue mentioned above, except that concurrency is allowed.

Compositionality plays an important role but needs to be modified, because 
gestures can last over any sequence of words in the sentence, and therefore there is no natural place for integrating the two modalities. Therefore, description of speech-gesture coordination cannot be given solely in a naive/static compositional way. Instead, as described in \cite{lawler2017gesture,rieser2017process}, speech and gesture processes operate in parallel to create initial representation for the meaning of the gesture and a partial semantic representation of the speech. Final meaning of the gesture is derived from the constraints present in the semantic representation of the speech. 
The latter is combined with the final meaning of the gesture \textit{compositionally}.

Notice that initially the two channels are \textit{observed} separately, and \textit{annotated}. This should remind us of the \texttt{\{I,O,B\}} annotations. In other words, we see a productive use of both compositional/inductive and process algebra/coinductive methods. 

Finally, we should mention that one of the first theoretical proposals to look at agent interaction as coinduction appeared in \cite{wegner1999coinductive}, and it included explicit mentions of NL dialogue and question answering. Within the following twenty years, as argued in in this article, the focus of NLP shifted towards coinductive methods, without ever mentioning the concept.\footnote{There are literally 12 entries mentioning "deep learning" and "coinduction", mostly accidentally, although  \cite{elliott2019generalized} (a formal analysis of convolutions) 
is an exception}

\subsubsection{Relating process algebras and coinduction}
Intuitively, process algebras are connected to coinduction as follows: 

--- A process algebra, from a very abstract level is a collection of processes, and can be viewed as a very complex automaton;

--- As we have above in Section \ref{sec:autascoalg}, an automaton can be viewed as a coalgebra;

\begin{figure}
\centering
\includegraphics[scale=0.35]{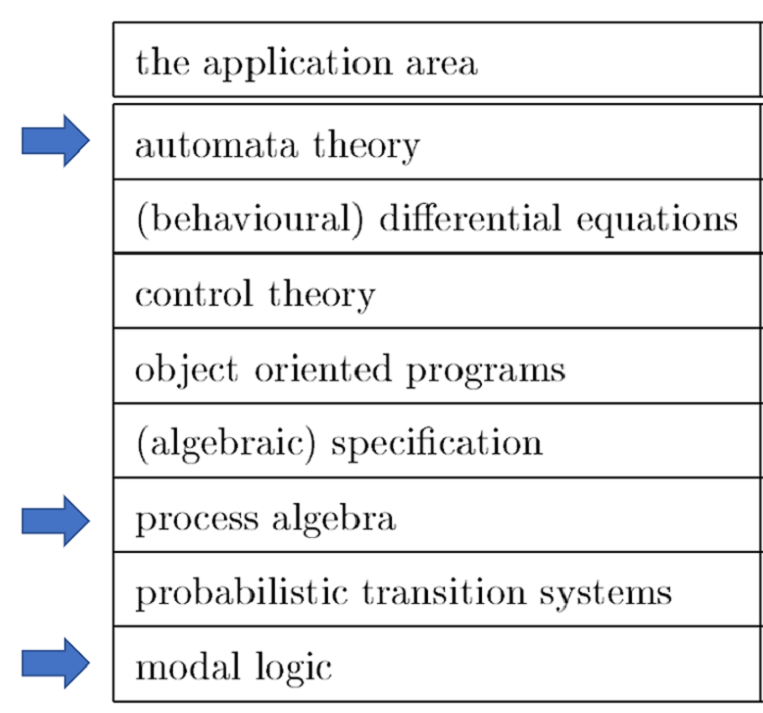}
\caption{Application areas for coalgebras, based on a table in \cite{kurz2001coalgebras}. The arrows, added by us, point to the areas directly applicable to computational linguistics. }\label{fig:kurztable}
\end{figure}

\FloatBarrier

The tutorial \cite{kurz2001coalgebras} discusses this relationship in details; 
see also \cite{ribeiro2006generic} for  ``a coinductive
rephrasal of classic process algebra."
Based on the cited sources, we can say: 

\begin{center}
Process algebra $ \simeq $ Coalgebra $ \simeq $ Coinduction 
\end{center}

That is, we can map between the three formalizations, even if such mappings could involve some subtle points. Informally, the formalisms can be viewed as roughly equivalent.
This can also be seen in Figure \ref{fig:kurztable}, based on \cite{kurz2001coalgebras}, showing application areas of the coalgebra approach (we added the arrows to point to the areas relevant for NLP). 

\color{black}

\section{NL understanding: coinduction and induction together}\label{sec:colaexam}

The hypothesis we are pursuing in this paper is that coinduction and induction are both needed for NL understanding. The argument we are making is based on computational and linguistic intuitions, and on circumstantial evidence. We would view the hypothesis as experimentally proven if NL models employing both approaches outperform other models for most tasks. As shown
in Section \ref{sec:togetherExs}, indeed there are relatively strong empirical arguments that this might be the case, although of course our selection of examples is biased.
Before we look at the empirical results, let's start with a manual analysis to develop an intuition for their joint use.

\subsection{A motivating example}\label{togetherMotEx}
In this section we want to show how coinduction and induction \textit{could} work together in assigning meaning to a poorly formed sentence. This is a fictional example, perhaps beyond capabilities of real systems. In the subsequent subsection, we will discuss implemented systems that combine neural and grammatical information. 

\bigskip
\begin{figure}[h]
\centering
\includegraphics[scale=0.42]{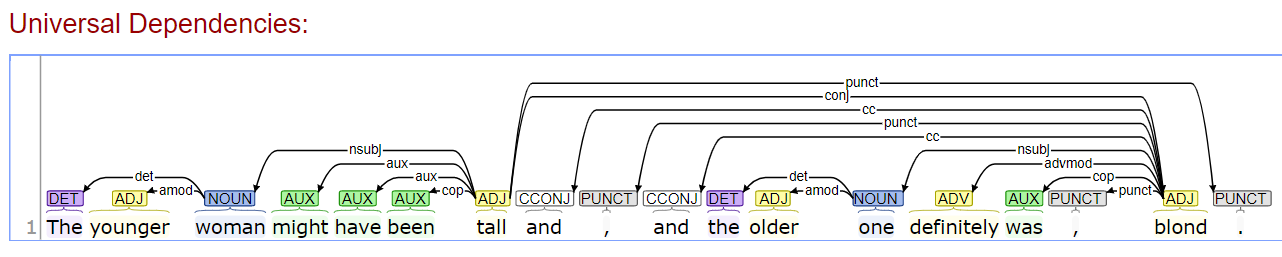}
\end{figure}

\begin{example}\label{ex:colasent}

\noindent
{Understanding a sentence from The Corpus of Linguistic Acceptability (CoLA)}\footnote{https://nyu-mll.github.io/CoLA/}
\citep{warstadt2018neural}. Consider one of the longer sentences from CoLA: 

\bs
\textit{
*The younger woman might have been tall and, and the older one definitely was, blond. }

Let us assume the sentence comes from a dialogue transcript. Dialogue sentences often have surprising structures.\footnote{A well known example is ``John thinks bananas."  The sentence is `ungrammatical' without the preceding ``What will Jane have for breakfast?"}
Even though marked ``ungrammatical" the sentence is understandable, and parses (perfectly?) with Stanford Stanza v.1.0.0 dependency parser\footnote{http://stanza.run/ } \citep{qi2020stanza}. However, the parse only has relationships between pairs words, so an interpretation of the sentence needs to be constructed. 
{The construction story, as we are imagining it, would be of coinductive and inductive composition, shown in the steps 1-6 below.}
 We can imagine the following sequence leading to an interpretation. For the sake of the argument, let us assume the interpretation will be a discourse representation structure 
(DRS).\footnote{\url{https://plato.stanford.edu/entries/discourse-representation-theory/}}

\begin{enumerate}
\item  Coinductively we find the boundaries of the sentence and sentence chunks.

Note.  Pattern matching works well here, e.g. probabilistic automata, or equivalent hidden Markov models for chunking; another option being 
 a combination of pattern matching and a neural network \cite{qi2020stanza}.
 
\item  Inductively (CFG) or coinductively (data trained dependency parser), structures are assigned to the parts and/or the whole. 

\item Some \texttt{predicate(argument)} semantics is assigned to the structures using pattern matching or attributed grammars (e.g. \cite{nltkbook} \cite{mccord2012deep}. This produces constraints on a resulting DRS.

Note. The method for assigning the constraints is not important; the point is that we do not have a full specification of semantics, but perhaps we have a partial one. 

\item  Coinductively (in written text), we find the “\texttt{, phrase,}” pattern (or construction) of a string between two commas. 

  
\item  Inductively, we try to apply the \textit{ellipsis} as a possible meaning frame.

Note: Actually several frames could be applied in parallel, e.g. apposition. 

\item  Inductively, we see that  \ \texttt{younger\_woman:tall and blond} \ is a possible interpretation. 
We add it to the DRS (and check its consistency).

\end{enumerate}

\end{example}

The main point of this example is to show the plausibility of using both pattern recognition (Steps 1, 3 and 4) and inductive steps (2, 5 and 6). Specifically, the possible meanings of ellipsis are usually given by a grammar.\footnote{
We also note that the process described above seems roughly consistent with the chunk-and-pass mechanism postulated by \cite{christiansen2016now}. In view of that mechanism, 
the ungrammaticality of the sentence could be attributed to the need to backtrack to apply the ellipsis, since the grammar does not license the ellipsis in the middle of the phrase.
An alternative interpretation would give us a repair, corresponding to \textit{The younger woman might have been tall and the older one definitely was blond} (a less interesting possibility, although more plausible if the sentence was spoken, with commas standing for pauses).}

\subsection{Recent examples of rules (induction) and neural networks (coinduction) working together}\label{sec:togetherExs}
In this section we present a few selected results in various subdomains of natural language processing showing superior performance of a combination of inductive and coinductive methods. We interpret them as supporting our view that induction and coinduction should both be used in the process of language understanding. Obviously, our selection is biased, but the examples are worth noting.  In the next section, we discuss examples which seem to be uniquely hard for NN, but are extremely easily formalized inductively; namely, we will switch to the world of mathematics.

\begin{example}\textbf{Predicate argument structure.}
Our first example, from \cite{he2017deep}, is a classical task of semantic role labeling (SRL) which is determining 
the predicate-argument structure of a sentence, i.e. 
“who did what to
whom.” The authors also discuss the impact of using ``gold syntax" (hand-annotated text data) and syntactic parsers on SRL. Their best model uses ``gold syntax." 

They observe the following relation between deep learning models and parsers: 
\begin{quote}
 Extensive empirical analysis of these
gains show that (1) deep models excel at
recovering long-distance dependencies but
can still make surprisingly obvious errors,
and (2) there is still room for syntactic
parsers to improve these results.
\end{quote}

However, and similarly to Fig. \ref{fig:claspSent}, they observe a quick drop in performance for longer argument dependencies. This is shown in Figure \ref{fig:heSRL}.

\begin{figure}
\centering
\includegraphics[scale=0.5]{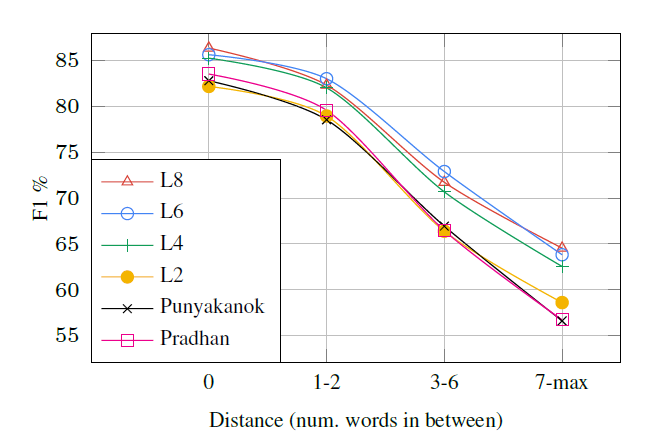}
\caption{Like earlier, in Fig.\ref{fig:claspSent}, we see a quick drop in performance with longer sentences, when dependencies become longer distance. The task is semantic role labeling and this figure is taken from the cited paper \cite{he2017deep}. }
\label{fig:heSRL} 
\end{figure}

\FloatBarrier
Their result and cited analysis can be viewed as supporting our thesis on the potential superiority of hybrid models. 
The cited paper uses 
a deep learning model (coinductive, by Section \ref{sec:autascoalg}) and an inductive complement, namely, a syntactic parser for encoding structure and the raw ``gold syntax" (from hand annotated data). When discussing the role of syntax in SRL, the article also says ``there
is significant room for improvement given oracle
syntax but errors from existing automatic parsers
prevent effective use in SRL." 

This suggests to us that more research is needed on how to effectively combine syntax and deep learning methods. Although the performance of hybrid methods is superior, the ground level problem of understanding the structure and meaning of longer sentences has not been solved. Likely, a solution to this problem will require a subtle interaction between the knowledge encoded in inductive structures and the statistical patterns encoded in weights of the deep learning model. \\

\color{black}

\end{example}

\begin{example}
\textbf{Information extraction: } Our next example also focuses on semantics. 
A recent article by \cite{tseo2020information} addresses the problem of extracting criteria for clinical trials using a novel architecture shown in Fig.\ref{fig:FB}. It combines
attention-based conditional random fields 
for named entity recognition (NER),  word2vec embedding
clustering for named entity linking (NEL), a context free grammar and a knowledge base. This system achieves a state of the art performance. 

The point of this examples, and the similar ones cited in this article, is to show that in practice we often benefit from a combination of methods. The other reason is to show that these are not one-off examples. Instead, they suggest, there is potential to generalize this practice using the tools of theoretical computer science, i.e. induction and coinduction.

\begin{figure}
\centering
\includegraphics[scale=0.6]{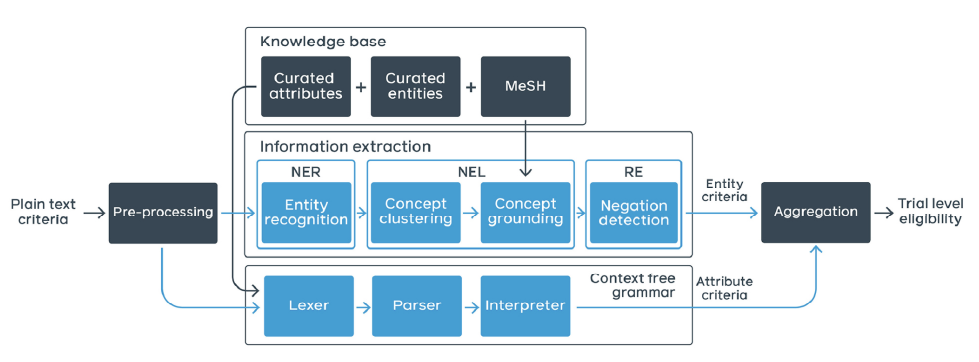}
\caption {\cite{tseo2020information} proposes another hybrid systems combining attention, word embeddings, context free grammars and several knowledge bases.}
\label{fig:FB}
\end{figure}

\end{example}

\normalsize

\color{black}
\FloatBarrier 

\begin{example} \textbf{Machine translation: comma analysis for rule-based machine translation and for patent translation.} 
Even though the two examples we cite below pertain to machine translation, they also touch upon two topics discussed earlier: the difficulty of parsing long sentences  and the complex structure of patents (Section \ref{sec:longsents}, as well as Fig.\ref{fig:cardinal} in Section \ref{sec:consvobs2}).
\color{black}

\begin{itemize}
\item Long sentences are not only difficult to parse and interpret, but also difficult to translate. In building an English to Korean machine translation system, \cite{kim2019comma} first classifies the usages of commas into nine standard grammatical functions, including various types of series (e.g. adjectives and conjuctions), parenthetical, appositives etc. This is done using a support vector machine. 
The  syntactic analysis is then performed according to the roles of the commas. The result of these preprocessing steps is improved quality of machine translation. 

\item The second example comes from the domain of Chinese-English patent machine translation.
\cite{li2016classifying} identify commas which separate
sub-sentences and non-sub-sentences, using two methods, one employing word knowledge
 and formal rules and the other machine learning. The rule-based method achieves over 93\% accuracy (F1) improves translation accuracy. As in the above examples, different functions of commas are discussed. 
\end{itemize}

\end{example}

\begin{example}\textbf{Parsing coinductively.} 

There is a tradition in linguistics to view syntax as a set of constraints, for example,  \cite{dalrymple1993lfg, frank1995principle}. This tradition is alive and well in more recent work, such as  
\cite{gotham2019glue}, who view syntax as constraining possible semantic interpretations. In a presentation,\footnote{We are citing almost verbatim from: \textit{Glue semantics for Universal Dependencies}, 
Matthew Gotham and Dag Haug, 
CLASP seminar, Gothenburg, 8 March 2018. The parentheticals and emphasis are ours.} they explain their reasons for viewing syntax as constraints:

\begin{itemize}
\item {What the approaches just mentioned }\textit{(i.e. Minimalism, Montague)} {have in common is the view that
syntactic structure plus lexical semantics determines interpretation.}

\item {From this it follows that if a sentence is ambiguous, \textit{(...)}, then
that ambiguity must be either lexical or syntactic.}  

\item {The Glue approach is that syntax \textbf{constrains} what can combine with
what, and how. }  
\end{itemize}

\end{example}

\noindent
In our view, these points also explain why constraint-driven approaches to semantics are less brittle. Constraints are not deterministic and they occasionally can be violated; the violations result in a penalty (i.e. higher value of the loss function), but without completely breaking of the parsing process. And although, bottom up parsing with a traditional, context free, grammar can be relatively robust and produce partial parses, the newer constraint-based parsers work better. 

For semantic dependency parsing, a good discussion and details of the constraint-based approach, used with three different dependency parsing formalisms, can be found in \cite{peng2017deep}, where constraints are applied to a decoder in a bidirectional LSTM neural network. 

Finally, we note the proposal for a more robust semantic parsing, based on a Type Theory with Records (TTR) \citep{cooper2005records}. Interestingly enough, the lemma {\textit{constrain(t)}} 
appears 36 times in the follow up work of \cite{dobnik2012modelling}. And we only need a small step to make it explicitly coinductive: 
TTR is based on a stratified type system, and is built bottom up to avoid the Russell paradox  \citep{cooper2012type}, but the stratification is already omitted in \cite{larsson2015formal} to ease the exposition. We know, from e.g. already cited \cite{barwise1996vicious}, that stratification can be replaced with constraints; therefore, purely constraint-based models of TTR should be consistent.

\begin{example}
\textbf{Morphology.} 

Our final example of the use of hybrid methods comes from a recent article by \cite{shmidman2020nakdan}. To quote:  
\begin{quote}
``The system combines
modern neural models with carefully curated declarative linguistic knowledge and
comprehensive manually constructed tables and dictionaries." 

``Our
approach (...) uses several bi-LSTM based
deep-learning modules for disambiguating
the correct diacritization in context. However, it
is also supplemented by comprehensive inflection
tables and lexicons, when appropriate."

\end{quote}

The problem the paper addresses is adding Hebrew diacritic markers. These markers are typically omitted in modern Hebrew, but have to be either implicitly understood or explicitly added to remove ambiguities. 
\end{example}

\noindent
\subsection{Comments and unresolved issues}\label{subsec:nlcoindcomments} 
We have presented examples where combinations of  methods improves performance on tasks ranging from semantics to morphology. 
The inductive parts varied, from facts, such as `gold syntax,' to tables and rules. The coinductive parts also varied. Clearly, much of the work described here is driven by experiments. And it seems that with enough effort and data almost any combination of architecture boxes can made to perform. What is missing, however, is our understanding: under what constraints, a combination of inductive and coinductive methods will achieve an optimum performance?

In other words, is there a principled way of building hybrid, inductive/coinductive or neuro-symbolic architectures?
Or, do such combinations of methods simply address deficiencies of specific neural networks, and with better NN systems, we can remove the inductive parts?  Again we have some anecdotal evidence that not everything should be done by coinduction, i.e. existing NN architectures show serious deficiencies for some tasks.

The question that arises seems to be: are these examples suggesting something deeper? In other words, is there a math underlying the practice? --- We know a few important mathematical facts about deep learning, starting with the classical result that neural networks can approximate any function. More recently, the superiority of deep neural networks over networks with only one hidden layer has been established  (e.g. \cite{telgarsky2016benefits}), and complementing it are results about the limitations of feed forward networks (e.g. \cite{mehrabi2018bounds}). These three facts suggest to us a technical problem: \textit{is there a formula balancing the inductive and coinductive contributions to understanding?} 

Perhaps this is the main question raised by the material discussed so far. We would like to know a formula quantifying the dependence of the accuracy of the neural network based on three factors: 1. the number of available hard coded facts/values; 2. amounts of available training data; and 3. difficulty of the problem.

Note that there are experimental results in this spirit. For example, \cite{sun2017revisiting} observe that performance improves logarithmically based on
volume of training data, which for some problems might be prohibitive. There is also interest in combining logic/rules with neural networks (see e.g. \cite{fischer2019dl2} and the next section). Recently, there appeared an analysis for a ``class
of hierarchically local compositional functions" \citep{poggio2020theoretical}, where it is hypothesized ``local hierarchical compositionality is imposed by
the wiring of our sensory cortex and, critically, is reflected in language."

However, there are also interesting questions possibly amenable to yes/no answers. For example, 
(1) Can a network with certain restriction on the learning function and parameters (depth, activation etc.) learn ten thousand natural language constructions of, say, depth-five from some fixed amount of data? -- This could perhaps be assessed using an artificial language approach in the spirit of \cite{baroni2012entailment} and  \cite{feinman2020learning}. An answer would perhaps shed some light on how difficult it is to create compositional semantics, given we have about ten thousand frequently used English verbs, each potentially having arguments and adjuncts.
(2) Same question, for long ``iterative" sentences with many sub-clauses,  such as appearing in patents and legal documents, as discussed in Section \ref{sec:longsents}.

Systems improve by learning from interaction, and we have provided argument for process-based, that is coinductive, approach to NL and multimodal dialogue. Having one formal view of different components of a NL understanding process should be helpful with integrating them. To us, it seems the best candidate for such formalism is coinduction/coalgebra. 

\color{black}

\section{Not everything should be done via coinduction}\label{sec:notEvery}

Given the success of deep learning models, the obvious question is whether ``everything should be done via coinduction", that is, using deep learning to find patterns in text, and create annotations representing the meaning of text. On the one hand, our common sense and intuitions contradict the idea of always doing pattern recognition from scratch, i.e. raw data. After all, there are databases of facts, many of them, e.g. financial and legal events, have been checked for accuracy. There are terabytes of texts containing useful information, including vocabulary definitions, and, finally, there are formal theories, created at great cost and effort, in mathematics, physics, biology etc.

Moreover, even though in theory, neural networks can approximate any function with arbitrary accuracy, in practice they show limitations. We will go over some of them using a few examples. So, this section can be viewed as complementing the previous one by adding `negative' examples.

\subsection{Compositionality is a challenge for neural networks}

As shown in \cite{zadrozny1994LF}, \textit{in theory,} any semantics can be encoded in a compositional fashion, and the encoding method is explicitly coinductive (via bisimulation). However, in practice, compositionality is a challenge for neural networks. Moreover, the lack of compositionality is an impediment to progress in NLP and cognitive modeling.

Our first set of observations come from M.Baroni's 2018 CLASP 
seminar.\footnote{\url{https://gu-clasp.github.io/static/fe1b398a82e63dd71c99e0668d706b79/1704969_marco-clasp-oct2018-composition.pdf}} and the follow up article \citep{baroni2020linguistic} His carefully designed experiment, using a simplified model of language understanding and trying to model language acquisition and compositionality, concludes with the following observations:  

\begin{itemize}
\item {(Recurrent) neural networks are remarkably powerful and general} (as agnostic ``end-to-end" learners from input-output pairs).
\item {They can generalize to new inputs that are different from those they
were trained on...}
\item {... but their generalization skills do not display systematic
compositionality}.
\item {Thus, they cannot adapt fast to continuous stream of new inputs in domains
such as language, math and, more generally, reasoning.}
\end{itemize}

\noindent
We can ask why this is the case that RNNs ``generalization skills do not display systematic compositionality"? --- An answer perhaps is that it is not their function. Most modern NN architectures incorporate sophisticated ways of computing correlations. In contrast, compositionality requires an association of a discrete construction with a specific meaning type, e.g. \cite{goldberg2006constructions} and \cite{cooper2015probabilistic}, 
and often access to background knowledge. 
In the process of understanding, the catalog of constructions is used to decompose a sentence into meaningful pieces (and the world knowledge helps with disambiguation). In principle such a catalog of constructions can be learned from data. 
But the practice is more complicated. Even in branches of mathematics, where the compositionality is trivially obvious to a human, neural networks have trouble learning it from data. 

Even on relatively simple, strictly compositional data sets, compositionality seems to elude standard NN architectures. After performing a detailed analysis of latent compositionality, from several natural points of view, \cite{hupkes2020compositionality} say ``high scores do still not necessarily imply that the trained models fully represent the true
underlying generative system."

\color{black} 
\subsection{Difficulties in learning algebra, arithmetic and formal reasoning}

\textbf{Algebra and arithmetic:} Even with special neural architectures, the performance on mathematical tasks is limited, and for general purpose NN (even out of the box transformers) it can be really disappointing \citep{saxton2019analysing, hupkes2020compositionality}.  \cite{madsen2020neural} attribute it to the  ``lack of inductive bias" in NN: 

``Neural networks can approximate complex functions, but they struggle to perform
exact arithmetic operations over real numbers. The lack of inductive bias for
arithmetic operations leaves neural networks without the underlying logic necessary
to extrapolate on tasks such as addition, subtraction, and multiplication."

The authors then proceed to create a special purpose architecture which provides better approximation of the three operations, but leave division as an open problem. \\

\noindent
\textbf{{Formal reasoning:}}  \cite{reimann2019neural} create a special purpose architecture, ``Neural Logic Rule Layers," to represent arbitrary logic rules in terms of their conjunctive and disjunctive normal forms. Their experiment shown relatively high accuracy (up to 98\% on a limited data set). The authors claim their approach might help create more explainable neural networks, as well as implement variants of fuzzy logic.
Slightly earlier, \cite{granmo2018tsetlin} proposed Tsetlin Machines for a similar task.

While these are interesting experiments, the question we want to ask is whether, from a practical point of view, hybrid systems, along the lines discussed in Section \ref{sec:togetherExs}, combining general purpose neural architectures with an inductive component, are better suited for building applications.

\subsection{Limitations of coinductive view of human cognition}

\cite{lake2015human} discussed the challenge of generalizing from a small number of examples, and introduced the Omniglot dataset for one-shot learning. Their progress report \citep{lake2019omniglot} concludes that ``recent approaches are still far from human-like concept learning on Omniglot, a challenge that requires performing many tasks with a single model."

In a related article, 
\cite{lake2017building} observe that deep neural networks have difficulties with tasks 
requiring common sense, causality and depth. The article gives an example of the difficulty neural networks have in generalizing in playing computer games by discussing  modified objectives, such as these: 
\begin{itemize}
\item Get the lowest possible score;
\item Get closest to 100, or any level, without going over;
\item Beat your friend, but just barely, not by too much;
\item Go as long as you can without dying;
\item Die as quickly as you can; and
\item Pass each level at the last possible minute.
\end{itemize}

\bs
The limitations of deep neural networks in modeling formal systems and human cognition suggest that further progress requires better accounts of grounding, compositionality and causality. To us, as the reader might already expect, it suggests the need for an interplay between the formal reasoning (inductive) and the data driven (coinductive) systems of reasoning.

\section{Discussion and summary}\label{sec:summary}

This article tries to reconcile two clashing paradigms by observing that a combination of inductive and coinductive methods produces superior results in many NLP tasks. It casts the observed empirical results within a common mathematical framework, a framework that has not been used so far in NL understanding, even though it has proven useful in other branches of computer science. This obviously leads to many questions of practical and theoretical importance. For example, we can ask how 
chunking and other cognitive phenomena fit into this formalized picture. And what about reasoning, generalization and abduction? However, in our view, the two key and most obvious concerns are:

\begin{enumerate}

\item \textbf{Practical importance:} { The \textit{main question} is whether having a common conceptual framework of coinduction + induction leads to better practical results, or better theoretical models of language understanding? }

--- We hope so, and this is a reason for writing this article. Perhaps an intuition can come from another, very different context. In the works on the traveling salesman and related NP-complete problems, it has been observed \citep{monasson1999determining} ``many NP-complete problems occurring in practice contain a mix of tractable and intractable constraints." This parallels the examples cited in this article that have successfully combined inductive and coinductive methods.

\item \textbf{Theoretical justification:} We have argued from empirical evidence that a combination of inductive and coinductive methods often produces superior results. Is there theoretical justification supporting these observations? For example, we have a theory proving that deeper neural networks produce better accuracy per number of neurons (Section \ref{subsec:nlcoindcomments}), if vanishing gradients can be controlled. In the more mathematical domain, it has been shown that 
NP-complete problems mentioned above, such as the traveling salesman problem’ and the boolean satisfiability (SAT) exhibit \textit{phase transitions }(cf.  \cite{cheeseman1991really,gent1994sat,monasson1999determining}).
Therefore, we ask whether there are ``phase transition" regions between applicability of inductive and coinductive methods.

\end{enumerate}

Interestingly enough, 
compositionality might be the key to progress in both issues. 
Traditionally, compositional semantics has taken its inspiration from mathematics, and has been specified by inductive definitions. This tradition is very strong in logic, linguistics and even computational linguistics. However, we see the emergence of 
compositionality in neural networks as a very active area of research. As of October 2020, per Google Scholar, we note about half of all the articles on deep learning and compositionality appeared in 2019 or later. We are seeing this increased interest because compositional systems are more interpretable, and can incorporate domain knowledge and known causal relationships. But since the world, knowledge and data are always changing, machine learning has to be part of the picture. We discussed in Section \ref{sec:notEvery} some of the limitations of the purely data driven attempts to derive compositionality from data, and earlier in Section \ref{sec:colaexam} empirical advantages of combining the two. \\

\noindent
\textbf{Summary:    } 
Coinduction is a mathematical and computational tool for specifying constraints on program behavior and data.  It is a natural complement to standard (i.e. bottom up) ways of defining compositional semantics or syntactic correctness. It provides a principled (formal) view of the current practice in human-computer interaction, parsing, machine translation and others. It covers possible worlds semantics, and, in theory, can encode any semantics. 
However, coinductive methods have practical limitations, as partial observations do not resolve all questions about interpretations, even though such methods are less brittle for longer texts.

In this article, we argued that a combination of inductive (e.g. traditional semantics) and coinductive (e.g. deep learning) methods is showing practical promise, and both approaches belong to the same computational paradigm. Perhaps now is the time for an explicit introduction of the term `coinduction' to the vocabulary of NLP.\\
\color{black}

\noindent
\textbf{Acknowledgments:}
The impetus for writing up these ideas about the potential of coinduction as a theoretical framework for natural language understanding came from a discussion with CLASP researchers during IWCS 2019 (although coinduction  guided some of my work work on dialogue systems in the second half of 1990s). This article organizes and expands on the main themes of the CLASP seminar talk given in May of 2020. 
In particular, I benefited from personal communications and comments of  S. Lappin, L. Moss, N. Ruozzi, N. Correa, R. Cooper, M. Steedman, and others. Thanks are due to O. Rambow for permission to reproduce Fig. 3;  to the authors of \cite{abzianidze2020first} for Fig. \ref{fig:claspSent}; to the authors of \cite{he2017deep} for  Fig.5; and to the authors of \cite{tseo2020information} for Fig. \ref{fig:FB}. 
Obviously, all the faults of this article are mine.

\color{black}

\color{black}
\small
\bibliographystyle{abbrvnat}
\bibliography{../../../../allRefCleaned2020}

\end{document}